\documentclass{article}

\usepackage{graphicx} % more modern
\usepackage{subfigure} 

\usepackage{natbib}

\usepackage{algorithm}
\usepackage{algorithmic}
\usepackage{amsmath}
\usepackage{amssymb}
\usepackage{hyperref}

\DeclareMathOperator*{\mean}{mean}

\usepackage[accepted]{icml2011}

\icmltitlerunning{Kernel PCA and its Applications}

\newcommand{\trans}{\mathrm{T}}
\newcommand{\ud}{\mathrm{d}}

\begin{document} 

\twocolumn[
\icmltitle{Kernel Principal Component Analysis and its Applications in \\ 
           Face Recognition and Active Shape Models}

\icmlauthor{Quan Wang}{wangq10@rpi.edu}
\icmladdress{Rensselaer Polytechnic Institute,
            110 Eighth Street, Troy, NY 12180 USA}
\icmlkeywords{Kernel PCA, Feature Extraction, Face Recognition, Active Shape Models}

\vskip 0.3in
]
%%%%%%%%%%%%%%%%%%%%%%%%%%%%%%%%%%%%%%%%%%%%%%%%%%%%%%%%%
%%%%%%%%%%%%%%%%%%%%%%%%%%%%%%%%%%%%%%%%%%%%%%%%%%%%%%%%%
%%%%%%%%%%%%%%%%%%%%%%%%%%%%%%%%%%%%%%%%%%%%%%%%%%%%%%%%%
\begin{abstract} 
Principal component analysis (PCA) is a popular tool for linear dimensionality reduction and feature extraction. Kernel PCA is the nonlinear form of PCA, which better exploits the complicated spatial structure of high-dimensional features. In this paper, we first review the basic ideas of PCA and kernel PCA. Then we focus on the reconstruction of pre-images for kernel PCA. We also give an introduction on how PCA is used in active shape models (ASMs), and discuss how kernel PCA can be applied to improve traditional ASMs. Then we show some experimental results to compare the performance of kernel PCA and standard PCA for classification problems. We also implement the kernel PCA-based ASMs, and use it to construct human face models. 
\end{abstract}

%%%%%%%%%%%%%%%%%%%%%%%%%%%%%%%%%%%%%%%%%%%%%%%%%%%%%%%%%
%%%%%%%%%%%%%%%%%%%%%%%%%%%%%%%%%%%%%%%%%%%%%%%%%%%%%%%%%
%%%%%%%%%%%%%%%%%%%%%%%%%%%%%%%%%%%%%%%%%%%%%%%%%%%%%%%%%
\section{Introduction}
\label{introduction}
In this section, we briefly review the principal component analysis method and the active shape models. 
%%%%%%%%%%%%%%%%%%%%%%%%%%%%%%%%%%%%%%%%%%%%%%%%%%%%%%%%%
%%%%%%%%%%%%%%%%%%%%%%%%%%%%%%%%%%%%%%%%%%%%%%%%%%%%%%%%%
%%%%%%%%%%%%%%%%%%%%%%%%%%%%%%%%%%%%%%%%%%%%%%%%%%%%%%%%%
\subsection{Principal Component Analysis}

Principal component analysis, or PCA, is a very popular technique for dimensionality reduction and feature extraction. PCA attempts to find a linear subspace of lower dimensionality than the original feature space, where the new features have the largest variance \cite{PRbook}. 

Consider a dataset $\{\mathbf{x}_i\}$ where $i=1,2,\cdots,N$, and each $\mathbf{x}_i$ is a $D$-dimensional vector. Now we want to project the data onto an $M$-dimensional subspace, where $M<D$. We assume the projection is denoted as $\mathbf{y=Ax}$, where $\mathbf{A}=[\mathbf{u}_1^\trans,\cdots,\mathbf{u}_M^\trans]$, and $\mathbf{u}_k^\trans\mathbf{u}_k=1$ for $k=1,2,\cdots,M$. We want to maximize the variance of $\{\mathbf{y}_i\}$, which is  the trace of the covariance matrix of $\{\mathbf{y}_i\}$. Thus, we want to find 
\begin{equation}
\mathbf{A}^*=\arg\max\limits_{\mathbf{A}} \mathrm{tr}(\mathbf{S_y}) , 
\end{equation}
where
\begin{equation}
\mathbf{S_y} = \frac{1}{N} \sum \limits_{i=1}^N 
(\mathbf{y}_i-\bar{\mathbf{y}})(\mathbf{y}_i-\bar{\mathbf{y}}) ^\trans , 
\end{equation}
and
\begin{equation}
\bar{\mathbf{y}}=\frac{1}{N}\sum\limits_{i=1}^N \mathbf{x}_i . 
\end{equation}
Let $ \mathbf{S_x} $ be the covariance matrix of $ \{\mathbf{x}_i\}$. 
Since $ \mathrm{tr}(\mathbf{S_y}) = \mathrm{tr}(\mathbf{AS_xA^\trans})$, by using the Lagrangian multiplier and taking the derivative, we get
\begin{equation}
\mathbf{S_xu}_k=\lambda_k\mathbf{u}_k, 
\end{equation}
which means that $ \mathbf{u}_k $ is an eigenvector of $\mathbf{S_x} $. 
Now $ \mathbf{x}_i $ can be represented as 
\begin{equation}
\mathbf{x}_i=\sum\limits_{k=1}^D \left( \mathbf{x}_i^\trans \mathbf{u}_k \right) \mathbf{u}_k . 
\end{equation}
$ \mathbf{x}_i $ can be also approximated by 
\begin{equation}
\label{eq6}
\widetilde{\mathbf{x}}_i=\sum\limits_{k=1}^M \left( \mathbf{x}_i^\trans \mathbf{u}_k \right) \mathbf{u}_k , 
\end{equation}
where $ \mathbf{u}_k $ is the eigenvector of $\mathbf{S_x} $ corresponding to the $k$th largest eigenvalue.

%%%%%%%%%%%%%%%%%%%%%%%%%%%%%%%%%%%%%%%%%%%%%%%%%%%%%%%%%
%%%%%%%%%%%%%%%%%%%%%%%%%%%%%%%%%%%%%%%%%%%%%%%%%%%%%%%%%
%%%%%%%%%%%%%%%%%%%%%%%%%%%%%%%%%%%%%%%%%%%%%%%%%%%%%%%%%
\subsection{Active Shape Models}

The active shape model, or ASM, is one of the most popular top-down object fitting approaches. It is designed to represent the complicated deformation patterns of the object shape, and to locate the object in new images. ASMs use the point distribution model (PDM) to describe the shape \cite{ASMs}. 
If a shape consists of $n$ points, and $(x_j,y_j)$ denotes the coordinates of the $j$th point, then the shape can be represented as a $2n$-dimensional vector
\begin{equation}
\mathbf{x}=[x_1,y_1,\cdots,x_n,y_n]^\trans . 
\end{equation} 
To simplify the problem, we now assume that all shapes have already been aligned. Otherwise, a rotation by $\theta$, a scaling by $s$, and a translation by $\mathbf{t}$ should be applied to $\mathbf{x}$. Given $N$ aligned shapes as training data, the mean shape can be calculated by 
\begin{equation}
\bar{\mathbf{x}}=\frac{1}{N}\sum\limits_{i=1}^N \mathbf{x}_i . 
\end{equation}
For each shape $ \mathbf{x}_i$ in the training set, its deviation from the mean shape 
$ \bar{\mathbf{x}} $ is
\begin{equation}
\ud \mathbf{x}_i=\mathbf{x}_i-\bar{\mathbf{x}}  . 
\end{equation}
Then the $2n\times2n$ covariance matrix $\mathbf{S}$ can be calculated by 
\begin{equation}
\mathbf{S}=\frac{1}{N}\sum\limits_{i=1}^N \ud\mathbf{x}_i \ud \mathbf{x}_i ^\trans . 
\end{equation}
Now we perform PCA on $\mathbf{S}$: 
\begin{equation}
\mathbf{Sp}_k=\lambda_k\mathbf{p}_k , 
\end{equation}
where $ \mathbf{p}_k $ is the eigenvector of $\mathbf{S}$ corresponding to the $k$th largest eigenvalue $\lambda_k$, and
\begin{equation}
\mathbf{p}_k^\trans\mathbf{p}_k=1 . 
\end{equation}
Let $\mathbf{P}$ be the matrix of the first $t$ eigenvectors: 
\begin{equation}
\mathbf{P}=[\mathbf{p}_1, \mathbf{p}_2, \cdots , \mathbf{p}_t]. 
\end{equation}
Then we can approximate a shape in the training set by 
\begin{equation}
\mathbf{x}=\bar{\mathbf{x}}+\mathbf{Pb} , 
\end{equation}
where $\mathbf{b}=[ b_1 ,  b_2 ,  \cdots ,  b_t]^\trans$ is the vector of weights
for different deformation patterns. By varying the parameters $b_k$, we can generate new examples of the shape. We can also limit each $b_k$ to constrain the deformation patterns of the shape. Typical limits are
\begin{equation}
-3\sqrt{\lambda_k} \leq b_k \leq 3\sqrt{\lambda_k} , 
\end{equation}
where $k=1,2,\cdots,t$. 
Another important issue of ASMs is how to search for the shape in new images using point distribution models. This problem is beyond the scope of our paper, and here we only focus on the statistical model itself.

%%%%%%%%%%%%%%%%%%%%%%%%%%%%%%%%%%%%%%%%%%%%%%%%%%%%%%%%%
%%%%%%%%%%%%%%%%%%%%%%%%%%%%%%%%%%%%%%%%%%%%%%%%%%%%%%%%%
%%%%%%%%%%%%%%%%%%%%%%%%%%%%%%%%%%%%%%%%%%%%%%%%%%%%%%%%%
\section{Kernel PCA} 

Standard PCA only allows linear dimensionality reduction. However, if the data has more complicated structures which cannot be well represented in a linear subspace, standard PCA will not be very helpful. Fortunately, kernel PCA allows us to generalize standard PCA to nonlinear dimensionality reduction \cite{kPCA1}.

%%%%%%%%%%%%%%%%%%%%%%%%%%%%%%%%%%%%%%%%%%%%%%%%%%%%%%%%%
%%%%%%%%%%%%%%%%%%%%%%%%%%%%%%%%%%%%%%%%%%%%%%%%%%%%%%%%%
%%%%%%%%%%%%%%%%%%%%%%%%%%%%%%%%%%%%%%%%%%%%%%%%%%%%%%%%%
\subsection{Constructing the Kernel Matrix}

Assume we have a nonlinear transformation $\phi(\mathbf{x})$ from the original 
$D$-dimensional feature space to an $M$-dimensional feature space, 
where usually $M \gg D$. Then each data point $\mathbf{x}_i$ is projected to a point $\phi(\mathbf{x}_i)$. We can perform standard PCA in the new feature space, but this can be extremely costly and inefficient. Fortunately, we can use kernel methods to simplify the computation \cite{kPCA3}. 

First, we assume that the projected new features have zero mean: 
\begin{equation}
\dfrac{1}{N}\sum\limits_{i=1}^N \phi(\mathbf{x}_i)=\mathbf{0}. 
\end{equation}
The covariance matrix of the projected features is $M \times M$, calculated by 
\begin{equation}
\label{eq17}
\mathbf{C}=\frac{1}{N}\sum\limits_{i=1}^N \phi(\mathbf{x}_i) \phi(\mathbf{x}_i) ^\trans . 
\end{equation} 
Its eigenvalues and eigenvectors are given by
\begin{equation}
\label{eq18}
\mathbf{Cv}_k=\lambda_k\mathbf{v}_k , 
\end{equation}
where $k=1,2, \cdots ,M$. From Eq. (\ref{eq17}) and Eq. (\ref{eq18}), we have
\begin{equation}
\label{eq19}
\frac{1}{N}\sum\limits_{i=1}^N \phi(\mathbf{x}_i)
\{ \phi(\mathbf{x}_i) ^\trans \mathbf{v}_k \}
=\lambda_k\mathbf{v}_k , 
\end{equation}
which can be rewritten as
\begin{equation}
\label{eq20}
\mathbf{v}_k
=\sum\limits_{i=1}^N a_{ki} \phi(\mathbf{x}_i) . 
\end{equation}
Now by substituting $\mathbf{v}_k$ in Eq. (\ref{eq19}) with Eq.  (\ref{eq20}), we have
\begin{equation}
\label{eq21}
\frac{1}{N}\sum\limits_{i=1}^N \phi(\mathbf{x}_i) \phi(\mathbf{x}_i)^\trans
\sum\limits_{j=1}^N a_{kj} \phi ( \mathbf{x}_j)
=\lambda_k \sum\limits_{i=1}^N a_{ki} \phi ( \mathbf{x}_i) . 
\end{equation}
If we define the kernel function
\begin{equation}
\label{eq22}
\kappa(\mathbf{x}_i,\mathbf{x}_j)=\phi(\mathbf{x}_i)^\trans \phi (\mathbf{x}_j) , 
\end{equation}
and multiply both sides of Eq. (\ref{eq21}) by $ \phi(\mathbf{x}_l) ^\trans $, we have 
\begin{equation}
\label{eq23}
\frac{1}{N}\sum\limits_{i=1}^N \kappa(\mathbf{x}_l,\mathbf{x}_i)
\sum\limits_{j=1}^N a_{kj} \kappa(\mathbf{x}_i,\mathbf{x}_j)
=\lambda_k \sum\limits_{i=1}^N a_{ki} \kappa(\mathbf{x}_l,\mathbf{x}_i) . 
\end{equation}
We can use the matrix notation
\begin{equation}
\label{eq24}
\mathbf{K}^2\mathbf{a}_k=\lambda_k N \mathbf{Ka}_k , 
\end{equation}
where 
\begin{equation}
\label{eq:Knm}
\mathbf{K}_{i,j}=\kappa(\mathbf{x}_i,\mathbf{x}_j),
\end{equation} 
and $\mathbf{a}_k$ is the $N$-dimensional column vector of $a_{ki}$:
\begin{eqnarray}
\mathbf{a_k}=[ \; a_{k1} \; a_{k2} \; \cdots \; a_{kN} \; ]^\trans  .
\end{eqnarray}
$\mathbf{a}_k$ can be solved by 
\begin{equation}
\label{eq25}
\mathbf{K}\mathbf{a}_k=\lambda_k N \mathbf{a}_k , 
\end{equation}
and the resulting kernel principal components can be calculated using
\begin{equation}
\label{eq26}
y_k(\mathbf{x})=\phi(\mathbf{x})^\trans \mathbf{v}_k
=\sum\limits_{i=1}^N a_{ki} \kappa(\mathbf{x},\mathbf{x}_i) . 
\end{equation}

If the projected dataset $\lbrace \phi(\mathbf{x}_i) \rbrace$ does not have zero mean, we can use the Gram matrix $ \widetilde{\mathbf{K}} $ to substitute the kernel matrix $ \mathbf{K} $. The Gram matrix is given by 
\begin{equation}
\label{eq27}
\widetilde{\mathbf{K}} =
\mathbf{K-1}_N\mathbf{K-K1}_N+\mathbf{1}_N\mathbf{K1}_N , 
\end{equation}
where $\mathbf{1}_N$ is the $N\times N$ matrix with all elements equal to $1/N$
\cite{PRbook}. 

The power of kernel methods is that we do not have to compute $ \phi(\mathbf{x}_i) $ explicitly. 
We can directly construct the kernel matrix from the training data set $ \{ \mathbf{x}_i \} $ \cite{kPCA2}. Two commonly used kernels are the polynomial kernel
\begin{equation}
\kappa(\mathbf{x,y)=(x^\trans y})^d , 
\end{equation}
or
\begin{equation}
\kappa(\mathbf{x,y)=(x^\trans y}+c)^d , 
\end{equation}
where $c>0$ is a constant, and the Gaussian kernel
\begin{equation}
\label{eq30}
\kappa(\mathbf{x,y})=\exp \left(   
- \Vert   \mathbf{x-y}   \Vert ^2 / 2 \sigma ^2
  \right) 
\end{equation}
with parameter $\sigma$. 

The standard steps of kernel PCA dimensionality reduction can be summarized as: 
 \begin{enumerate}
\item Construct the kernel matrix $\mathbf{K}$ from the training data set 
$ \{ \mathbf{x}_i \} $ using Eq. (\ref{eq:Knm}). 
\item Compute the Gram matrix $\widetilde{\mathbf{K}}$ using Eq. (\ref{eq27}). 
\item Use Eq. (\ref{eq25}) to solve for the 
vectors $\mathbf{a}_i$ (substitute $\mathbf{K}$ with $\widetilde{\mathbf{K}}$).
\item Compute the kernel principal components $y_k(\mathbf{x})$ using Eq. (\ref{eq26}). 
 \end{enumerate}

%%%%%%%%%%%%%%%%%%%%%%%%%%%%%%%%%%%%%%%%%%%%%%%%%%%%%%%%%
%%%%%%%%%%%%%%%%%%%%%%%%%%%%%%%%%%%%%%%%%%%%%%%%%%%%%%%%%
%%%%%%%%%%%%%%%%%%%%%%%%%%%%%%%%%%%%%%%%%%%%%%%%%%%%%%%%%
\subsection{Reconstructing Pre-Images}
So far, we have discussed how to generate new features $y_k(\mathbf{x})$ using kernel PCA. This is enough for applications such as feature extraction and data classification. However, for some other applications, we need to approximately reconstruct the pre-images $\{\mathbf{x}_i\}$ from the kernel PCA features $\{\mathbf{y}_i\}$. This is the case in active shape models, where we not only need to use PCA features to describe the deformation patterns, but also have to reconstruct the shapes from the PCA features \cite{PCA_ASM1,PCA_ASM2}. 

In standard PCA, the pre-image $ \mathbf{x}_i $ can simply be approximated by Eq. (\ref{eq6}). However, Eq. (\ref{eq6}) cannot be used for kernel PCA \cite{pre_image2}. For kernel PCA, we define a projection operator $P_m$ which projects $\phi(\mathbf{x})$ to its approximation
\begin{equation}
P_m \phi(\mathbf{x}) =
\sum\limits _{k=1}^m y_k (\mathbf{x} ) \mathbf{v}_k , 
\end{equation}
where $ \mathbf{v}_k $ is the eigenvector of the $\mathbf{C}$ matrix, which is define by Eq. (\ref{eq17}). If $m$ is large enough, we have 
$ P_m \phi(\mathbf{x}) \approx \phi(\mathbf{x}) $. Since finding the exact pre-image $\mathbf{x}$ is difficult, we turn to find an approximation $\mathbf{z}$ such that 
\begin{equation}
\phi ( \mathbf{z})\approx P_m \phi(\mathbf{x}) . 
\end{equation}
This can be approximated by minimizing
\begin{equation}
\rho(\mathbf{z})=
\Vert            \phi ( \mathbf{z})- P_m \phi(\mathbf{x})        \Vert ^2 . 
\end{equation}

%%%%%%%%%%%%%%%%%%%%%%%%%%%%%%%%%%%%%%%%%%%%%%%%%%%%%%%%%
%%%%%%%%%%%%%%%%%%%%%%%%%%%%%%%%%%%%%%%%%%%%%%%%%%%%%%%%%
%%%%%%%%%%%%%%%%%%%%%%%%%%%%%%%%%%%%%%%%%%%%%%%%%%%%%%%%%
\subsection{Pre-Images for Gaussian Kernels}

There are some existing techniques to compute $\mathbf{z}$ for specific kernels \cite{pre_image1}. 
For a Gaussian kernel 
$\kappa(\mathbf{x,y})=\exp \left(   - \Vert   \mathbf{x-y}   \Vert ^2 / 2 \sigma ^2 \right)$, 
$\mathbf{z}$ should satisfy
\begin{equation}
\mathbf{z}=\frac
{  
\sum\limits_{i=1}^N \gamma_i 
\exp \left(   - \Vert   \mathbf{z}-\mathbf{x}_i   \Vert ^2 / 2 \sigma ^2 \right) \mathbf{x}_i 
}
{
\sum\limits_{i=1}^N \gamma_i 
\exp \left(   - \Vert   \mathbf{z}-\mathbf{x}_i   \Vert ^2 / 2 \sigma ^2 \right)
} , 
\end{equation}
where 
\begin{equation}
\gamma_i=\sum\limits_{k=1}^m y_ka_{ki} . 
\end{equation}
We can compute $\mathbf{z}$ iteratively: 
\begin{equation}
\label{eq36}
\mathbf{z}_{t+1}=\frac
{  
\sum\limits_{i=1}^N \gamma_i 
\exp \left(   - \Vert   \mathbf{z}_t-\mathbf{x}_i   \Vert ^2 / 2 \sigma ^2 \right) \mathbf{x}_i 
}
{
\sum\limits_{i=1}^N \gamma_i 
\exp \left(   - \Vert   \mathbf{z}_t-\mathbf{x}_i   \Vert ^2 / 2 \sigma ^2 \right)
} . 
\end{equation}

%%%%%%%%%%%%%%%%%%%%%%%%%%%%%%%%%%%%%%%%%%%%%%%%%%%%%%%%%
%%%%%%%%%%%%%%%%%%%%%%%%%%%%%%%%%%%%%%%%%%%%%%%%%%%%%%%%%
%%%%%%%%%%%%%%%%%%%%%%%%%%%%%%%%%%%%%%%%%%%%%%%%%%%%%%%%%
\section{Experiments}
In this section, we show the setup and results of our three experiments. 
The first two experiments are classification problems without 
pre-image reconstruction. The third experiment combines active shape 
models with kernel PCA, and involves the pre-image reconstruction algorithm. 

%%%%%%%%%%%%%%%%%%%%%%%%%%%%%%%%%%%%%%%%%%%%%%%%%%%%%%%%%
%%%%%%%%%%%%%%%%%%%%%%%%%%%%%%%%%%%%%%%%%%%%%%%%%%%%%%%%%
%%%%%%%%%%%%%%%%%%%%%%%%%%%%%%%%%%%%%%%%%%%%%%%%%%%%%%%%%
\subsection{Pattern Classification for Synthetic Data}

Before we work on real data, we would like to generate some synthetic datasets and test our algorithm on them. In this paper, we use the two-concentric-spheres data. 

\subsubsection{Data Description}

\begin{figure}[h!]
\vskip 0.2in
\begin{center}
\centerline{\includegraphics[width=\columnwidth]{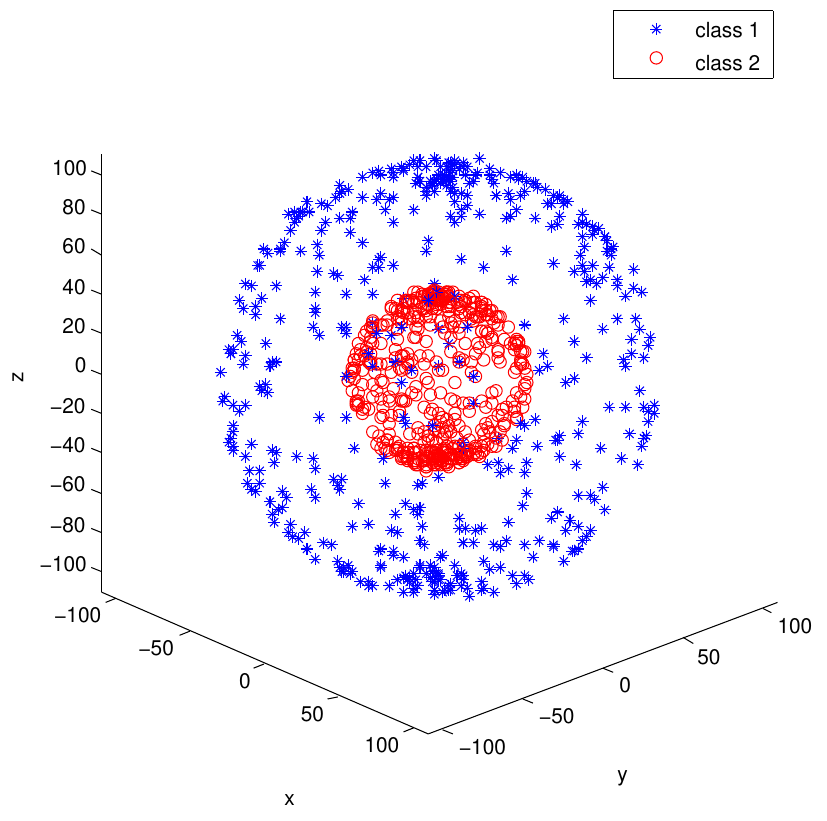}}
\caption{
3D plot of the two-concentric-spheres synthetic data.
}
\label{data3}
\end{center}
\vskip -0.2in
\end{figure}

We assume that we have equal numbers or data points distributed on two concentric sphere surfaces. 
If $N$ is the total number of all data points, then we have $N/2$  class 1 points on a sphere of radius $r_1$, and $N/2$ class 2 points on a sphere of radius $r_2$. In the spherical coordinate system, the inclination (polar angle) $\theta$ is uniformly distributed in $[0,\pi]$, and the azimuth (azimuthal angle) $ \phi $ is uniformly distributed in $ [0,2\pi) $ for both classes. Our observations of the data points are the $(x,y,z)$ coordinates in the Cartesian coordinate system, and all the three coordinates are perturbed by a Gaussian noise of standard deviation $\sigma_\mathrm{noise}$. 
We set $N=1000$, $r_1=40$, $r_2=100$, $\sigma_\mathrm{noise}=1$, and give a 3D plot of the data in Figure \ref{data3}. 

\subsubsection{PCA and Kernel PCA Results}

\begin{figure}[h!]
\vskip 0.2in
\begin{center}
\centerline{\includegraphics[width=\columnwidth]{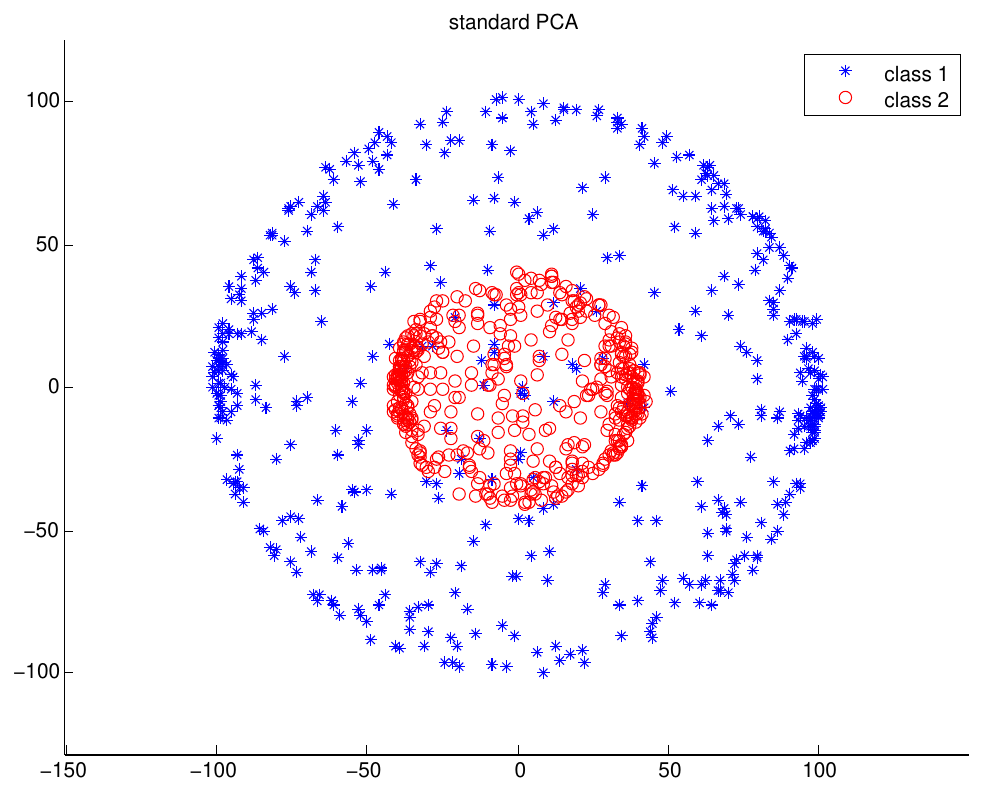}}
\caption{
Standard PCA results for the two-concentric-spheres synthetic data.
}
\label{data3_PCA}
\end{center}
\vskip -0.2in
\end{figure}

\begin{figure}[h!]
\vskip 0.2in
\begin{center}
\centerline{\includegraphics[width=\columnwidth]{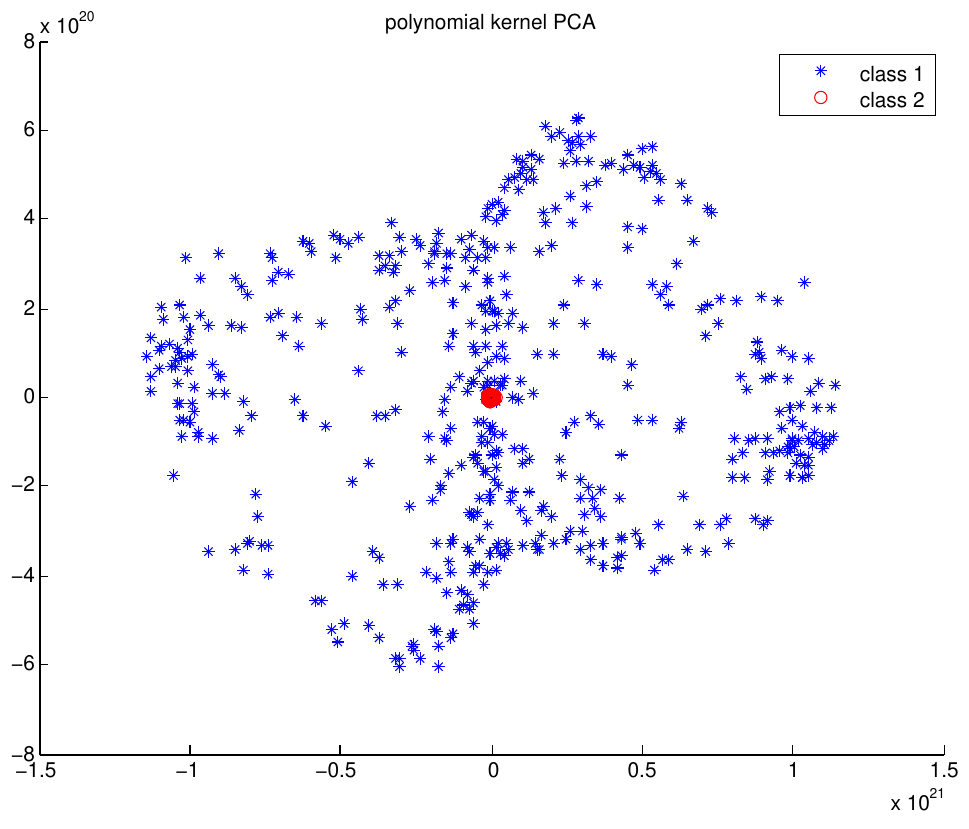}}
\caption{
Polynomial kernel PCA results for the two-concentric-spheres synthetic data with $d=5$.
}
\label{data3_kPCA_poly5}
\end{center}
\vskip -0.2in
\end{figure}

\begin{figure}[h!]
\vskip 0.2in
\begin{center}
\centerline{\includegraphics[width=\columnwidth]{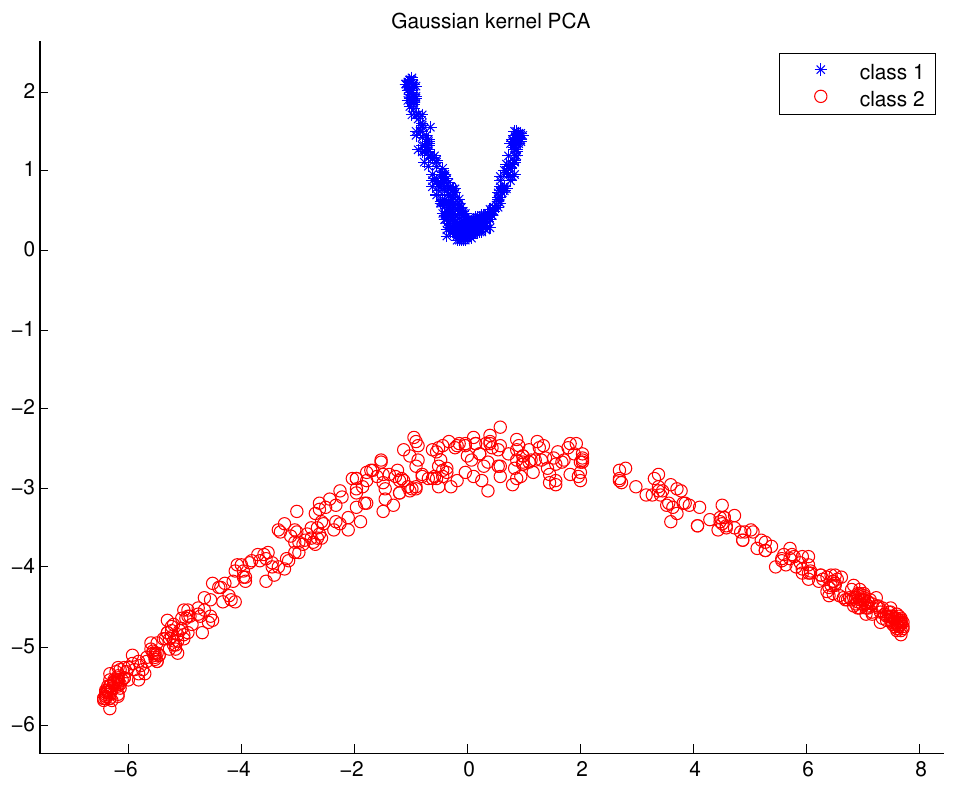}}
\caption{
Gaussian kernel PCA results for the two-concentric-spheres synthetic data with $\sigma=27.8$.
}
\label{data3_kPCA_gaussian}
\end{center}
\vskip -0.2in
\end{figure}

To visualize our results, we project the original 3-dimensional data onto a 2-dimensional feature space by using both standard PCA and kernel PCA. For kernel PCA, we use a polynomial kernel with $d=5$, and a Gaussian kernel with $\sigma=27.8$ (parameter selection is discussed in Section \ref{discussion}). The results of standard PCA, polynomial kernel PCA, and Gaussian kernel PCA are given in 
Figure \ref{data3_PCA},  Figure \ref{data3_kPCA_poly5}, and Figure \ref{data3_kPCA_gaussian}, respectively. 

We note that here though we mark points in different classes with different colors, we are actually doing unsupervised learning. Neither standard PCA nor kernel PCA takes the class labels as their input. 

In the resulting figures we can see that, standard PCA does not reveal any structural information of the original data. For polynomial kernel PCA, in the new feature space, class 2 data points are clustered while class 1 data points are scattered. But they are still not linearly separable. For Gaussian kernel PCA, the two classes are completely linearly separable.

%%%%%%%%%%%%%%%%%%%%%%%%%%%%%%%%%%%%%%%%%%%%%%%%%%%%%%%%%
%%%%%%%%%%%%%%%%%%%%%%%%%%%%%%%%%%%%%%%%%%%%%%%%%%%%%%%%%
%%%%%%%%%%%%%%%%%%%%%%%%%%%%%%%%%%%%%%%%%%%%%%%%%%%%%%%%%
\subsection{Classification for Aligned Human Face Images}

After we have tested our algorithm on synthetic data, we would like to use it for real data classification. 
Here we use PCA and kernel PCA to extract features from human face images, and use the simplest linear classifier \cite{linear} for classification. Then we compare the error rates of using PCA and kernel PCA. 

\subsubsection{Data Description}

\begin{figure}[h!]
\vskip 0.2in
\begin{center}
\centerline{\includegraphics[width=\columnwidth]{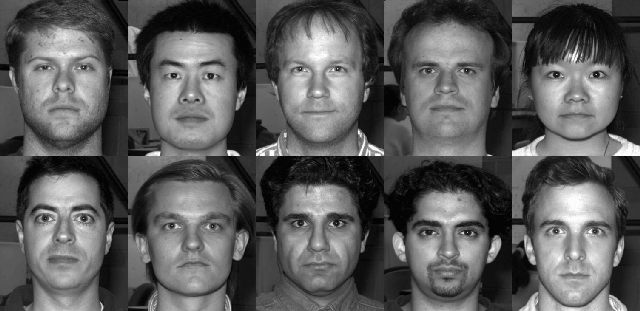}}
\caption{
Example images from the Yale Face Database B.
}
\label{yaleBfaces}
\end{center}
\vskip -0.2in
\end{figure}

For this task, we use images from the Yale Face Database B \cite{yale}, which contains 5760 single light source gray-level images of 10 subjects, each seen under 576 viewing conditions. We take 51 images of the first subject and  51 images of the third subject as the training data, and 13 images of each of them as testing data. Then all the images are aligned, and each image has $168 \times 192 $ pixels. 
Example images of the Yale Face Database B are shown in Figure \ref{yaleBfaces}. 

\subsubsection{Classification Results}

We use the $168 \times 192 $ pixel intensities as the original features for each image, thus the original feature vector is $32256$-dimensional. Then we use standard PCA and Gaussian kernel PCA to extract the 9 most significant features from the training data, and record the eigenvectors. 

For standard PCA, only the eigenvectors are needed to extract features from testing data. For Gaussian kernel PCA, both the eigenvectors and the training data are needed to extract features from testing data. Note that for standard PCA, there are particular fast algorithms to compute the eigenvectors when the dimensionality is much higher than the number of data points \cite{PRbook}. 

For kernel PCA, we use a Gaussian kernel with $\sigma=22546$ (we will talk about how to select the parameters in Section \ref{discussion}). For classification, we use the simplest linear classifier \cite{linear}. The training error rates and the testing error rates for standard PCA and Gaussian kernel PCA are given in Table \ref{errorRate}. We can see that Gaussian kernel PCA achieves much lower error rates than standard PCA.

\begin{table}[t]
\caption{ 
Classification error rates on training data and testing data for standard PCA and Gaussian kernel PCA with $\sigma=22546$.
}
\label{errorRate}
\vskip 0.15in
\begin{center}
\begin{small}
\begin{sc}
\begin{tabular}{|c|c|c|}
\hline
\abovespace\belowspace
 Error Rate & PCA & Kernel PCA \\
\hline
\abovespace
Training   Data  & 8.82\% & 6.86\%  \\
\belowspace
Testing Data & 23.08\% & 11.54\% \\
\hline
\end{tabular}
\end{sc}
\end{small}
\end{center}
\vskip -0.1in
\end{table}

%%%%%%%%%%%%%%%%%%%%%%%%%%%%%%%%%%%%%%%%%%%%%%%%%%%%%%%%%
%%%%%%%%%%%%%%%%%%%%%%%%%%%%%%%%%%%%%%%%%%%%%%%%%%%%%%%%%
%%%%%%%%%%%%%%%%%%%%%%%%%%%%%%%%%%%%%%%%%%%%%%%%%%%%%%%%%
\subsection{Kernel PCA-Based Active Shape Models}

In ASMs, the shape of an object is described with point distribution models, and standard PCA is used to extract the principal deformation patterns from the shape vectors
 $\{ \mathbf{x}_i \}$. If we use kernel PCA instead of standard PCA here, it is promising that we will be able to discover more hidden deformation patterns. 

\subsubsection{Data Description}

\begin{figure}[h!]
\vskip 0.2in
\begin{center}
\centerline{\includegraphics[width=0.5\columnwidth]{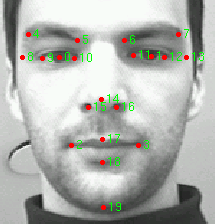}}
\caption{
The 20 manually annotated feature points on an image ($286 \times 384$) from BioID. 
}
\label{bioid}
\end{center}
\vskip -0.2in
\end{figure}

In our work, we use Tim Cootes' manually annotated points of 1521 human face images from the BioID database. For each face image, 20 feature points (landmarks) are annotated, as shown in Figure \ref{bioid}. Thus the original feature vector for each image is $40$-dimensional (two coordinates for each landmark).

\subsubsection{Experimental Results}

In our work, we first normalize all the shape vectors by restricting both the $x$ coordinates and $y$ coordinates in the range $[0,1]$. Then we perform standard PCA and Gaussian kernel PCA on the normalized shape vectors. For standard PCA, the reconstruction of the shape is given by 
\begin{equation}
\mathbf{x}=\bar{\mathbf{x}}+\mathbf{Pb} .
\end{equation}
For kernel PCA, the reconstruction of the shape is given by
\begin{equation}
\mathbf{z}=\Omega(\mathbf{y}) , 
\end{equation}
where $ \Omega(\mathbf{y}) $ denotes the reconstruction algorithm described by Eq. (\ref{eq36}). 

\begin{figure}[h!]
\vskip 0.2in
\begin{center}
\centerline{\includegraphics[width=\columnwidth]{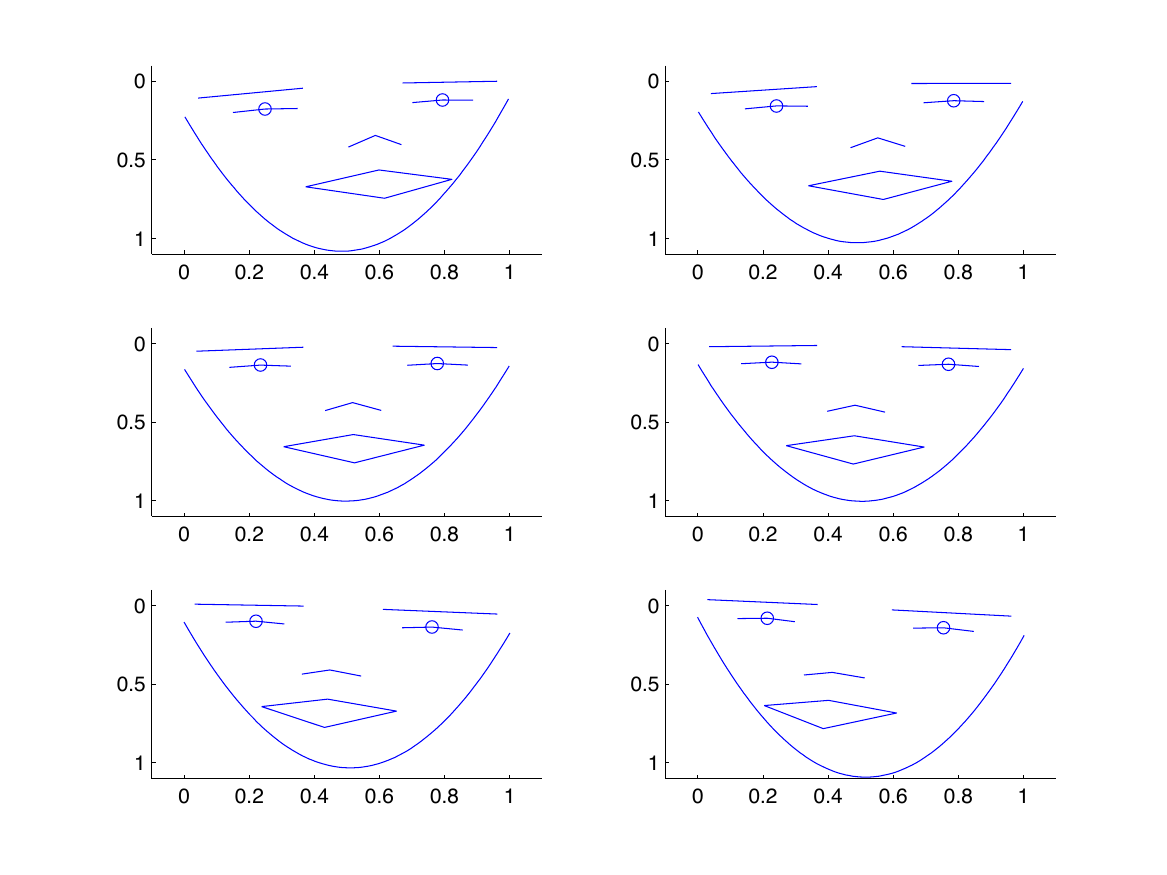}}
\caption{
The effect of varying the first PCA feature for ASM. 
}
\label{PCA_ASM_1}
\end{center}
\vskip -0.2in
\end{figure}

\begin{figure}[h!]
\vskip 0.2in
\begin{center}
\centerline{\includegraphics[width=\columnwidth]{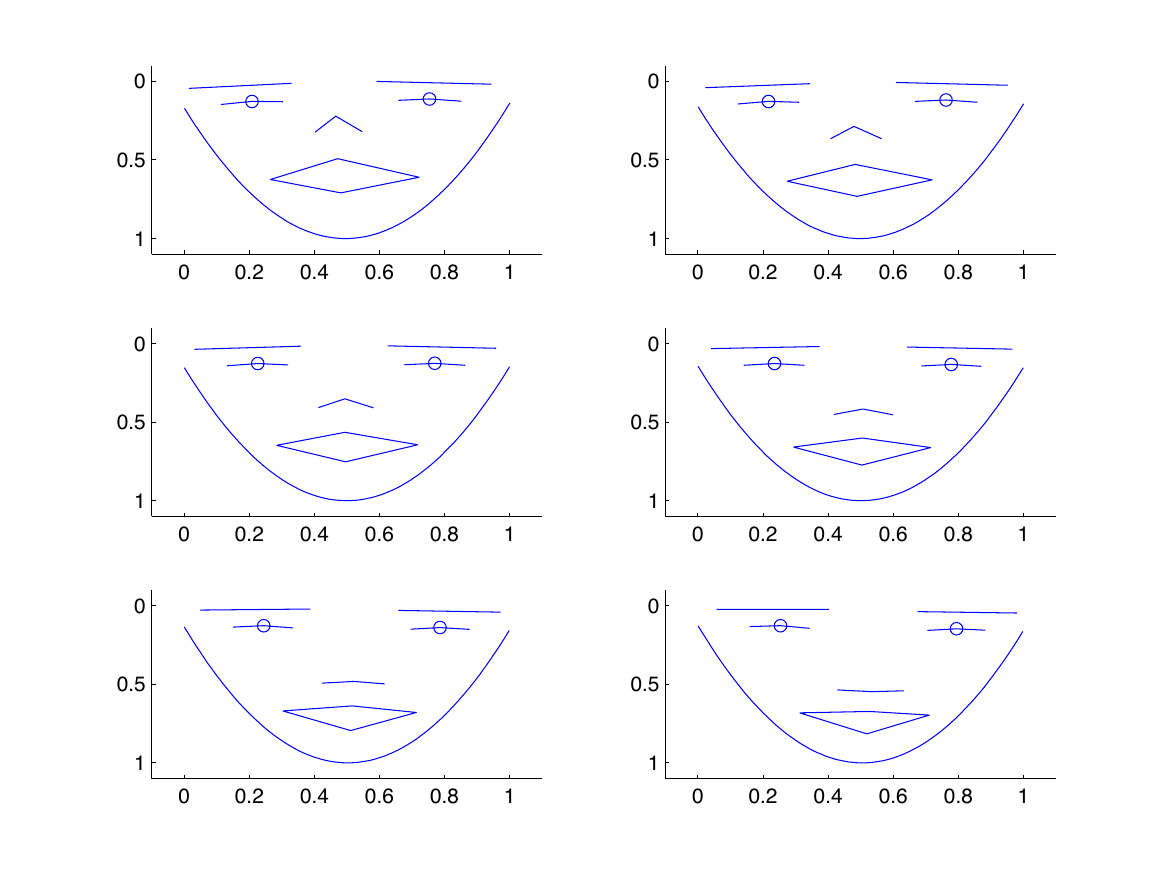}}
\caption{
The effect of varying the second PCA feature for ASM. 
}
\label{PCA_ASM_2}
\end{center}
\vskip -0.2in
\end{figure}

\begin{figure}[h!]
\vskip 0.2in
\begin{center}
\centerline{\includegraphics[width=\columnwidth]{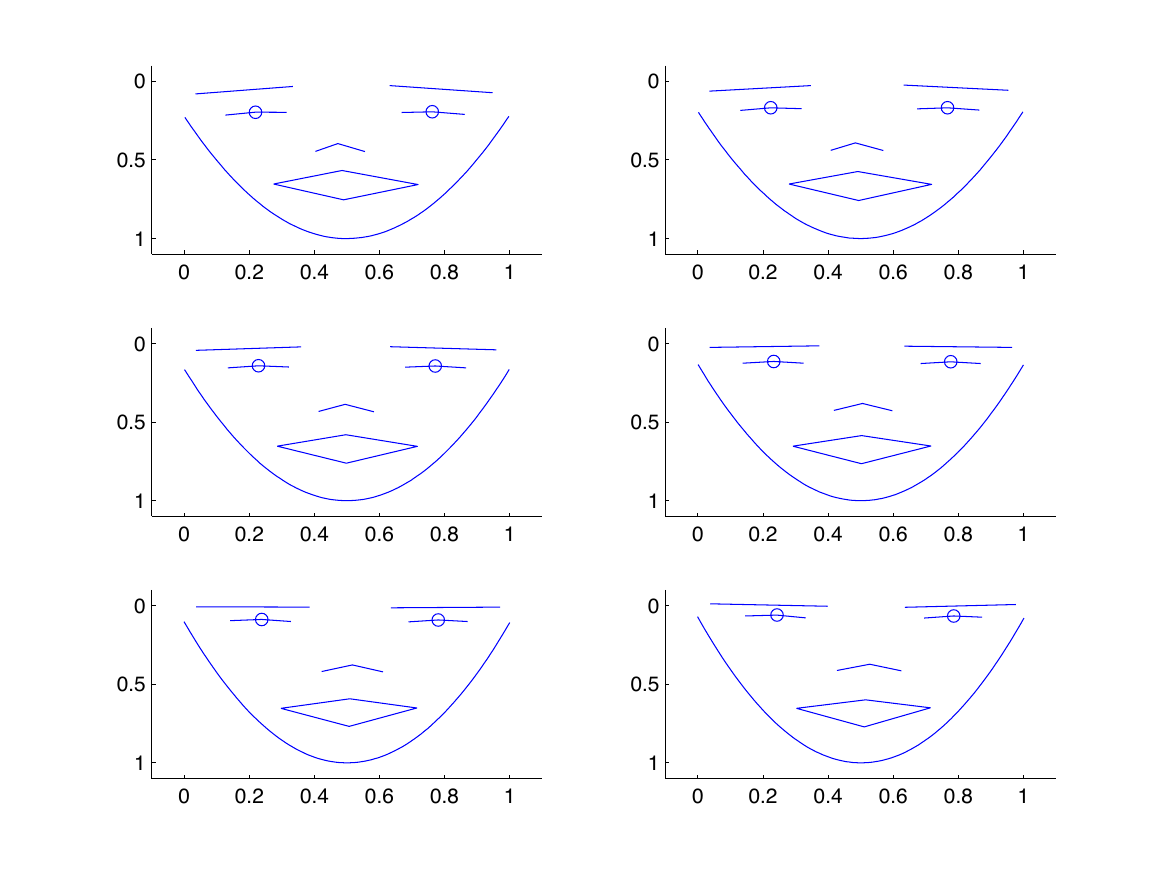}}
\caption{
The effect of varying the third PCA feature for ASM. 
}
\label{PCA_ASM_3}
\end{center}
\vskip -0.2in
\end{figure}

For standard PCA, we focus on studying the deformation pattern associated with each entry of $ \mathbf{b} $. That is to say, each time we uniformly select different values of $ b_k$ in 
$ [-3\sqrt{\lambda_k} , 3\sqrt{\lambda_k}] $, and set $b_{k'}=0$ for all $k'\neq k$. The effects of varying the first PCA feature, the second PCA feature, and the third PCA feature are shown in Figure \ref{PCA_ASM_1}, Figure \ref{PCA_ASM_2} and Figure \ref{PCA_ASM_3}, respectively. 
The face is drawn by using line segments to represent eye brows, eyes, the nose, using circles to represent eye balls, using a quadrilateral to represent the mouth, and fitting a parabola to represent the contour of the face.

\begin{figure}[h!]
\vskip 0.2in
\begin{center}
\centerline{\includegraphics[width=\columnwidth]{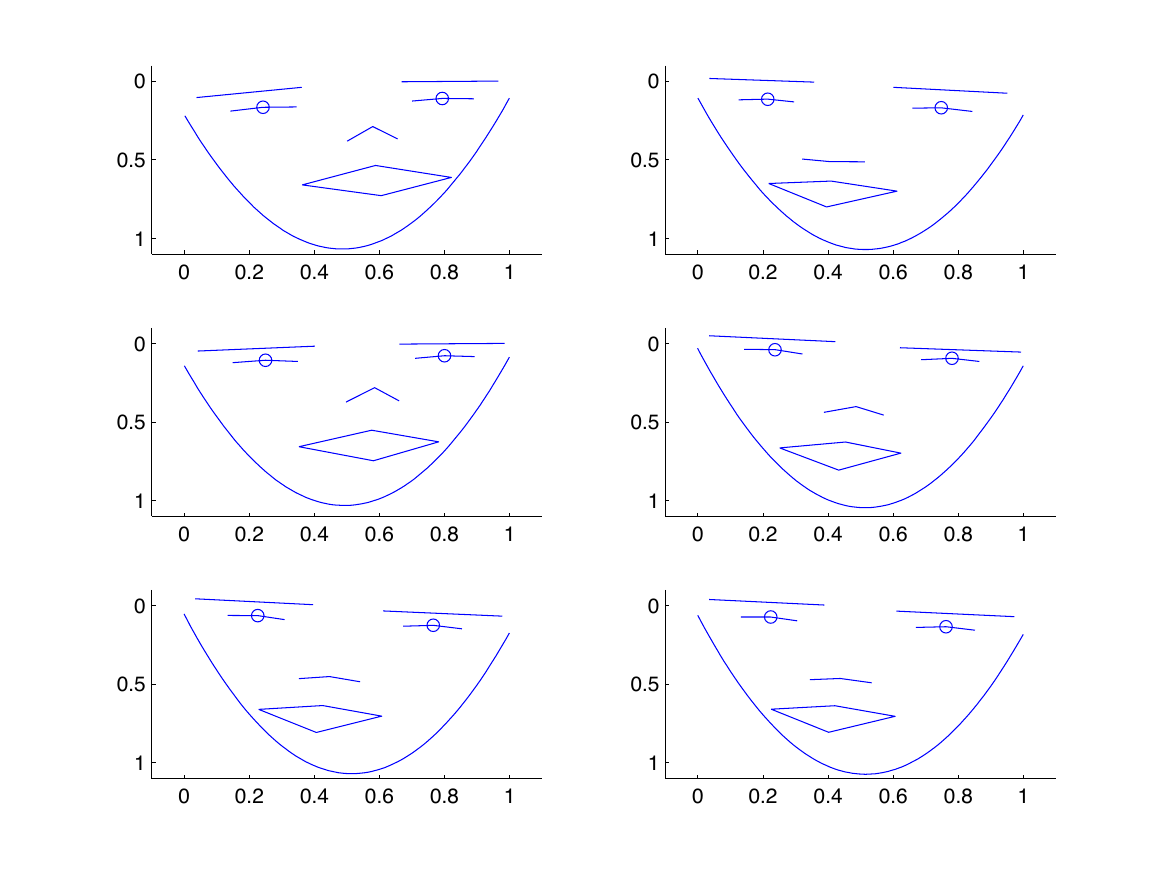}}
\caption{
The effect of varying the first  Gaussian kernel PCA feature for ASM. 
}
\label{kPCA_ASM_1}
\end{center}
\vskip -0.2in
\end{figure}

\begin{figure}[h!]
\vskip 0.2in
\begin{center}
\centerline{\includegraphics[width=\columnwidth]{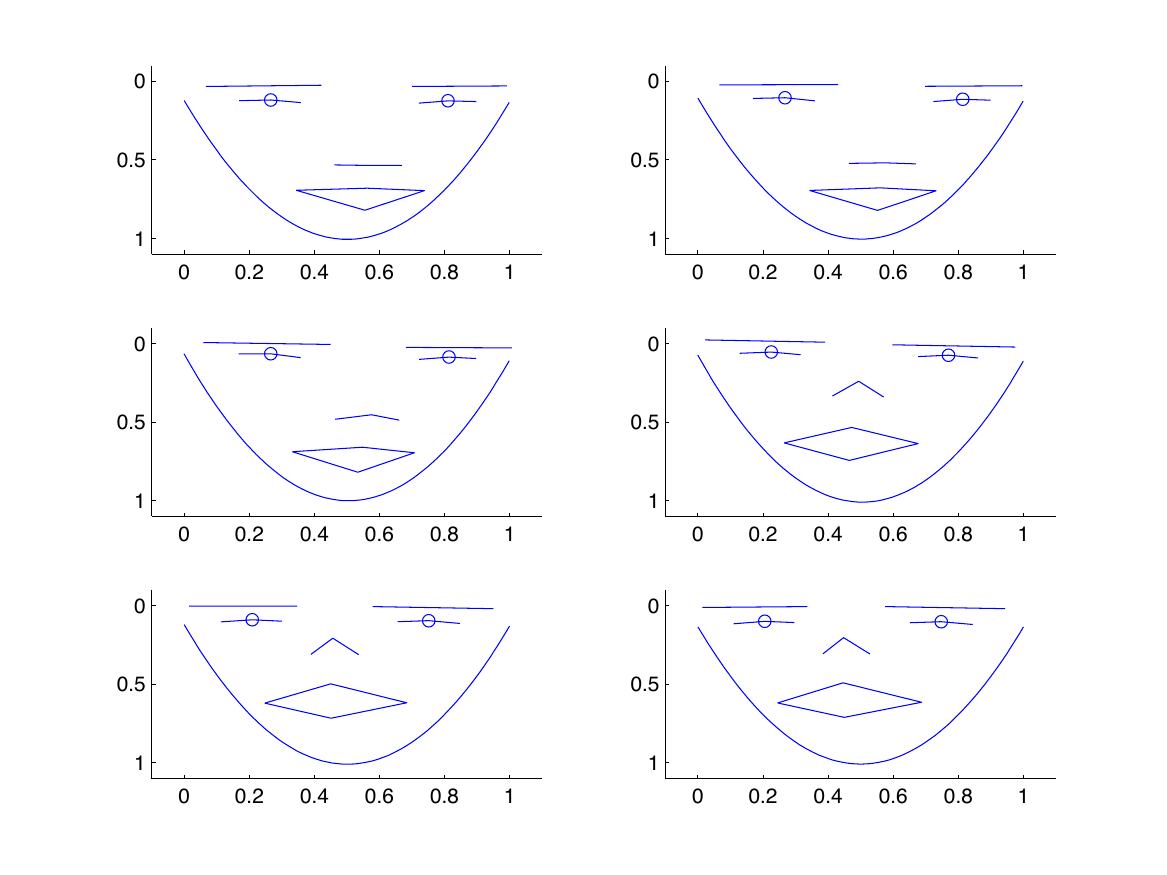}}
\caption{
The effect of varying the second Gaussian kernel PCA feature for ASM. 
}
\label{kPCA_ASM_2}
\end{center}
\vskip -0.2in
\end{figure}

\begin{figure}[h!]
\vskip 0.2in
\begin{center}
\centerline{\includegraphics[width=\columnwidth]{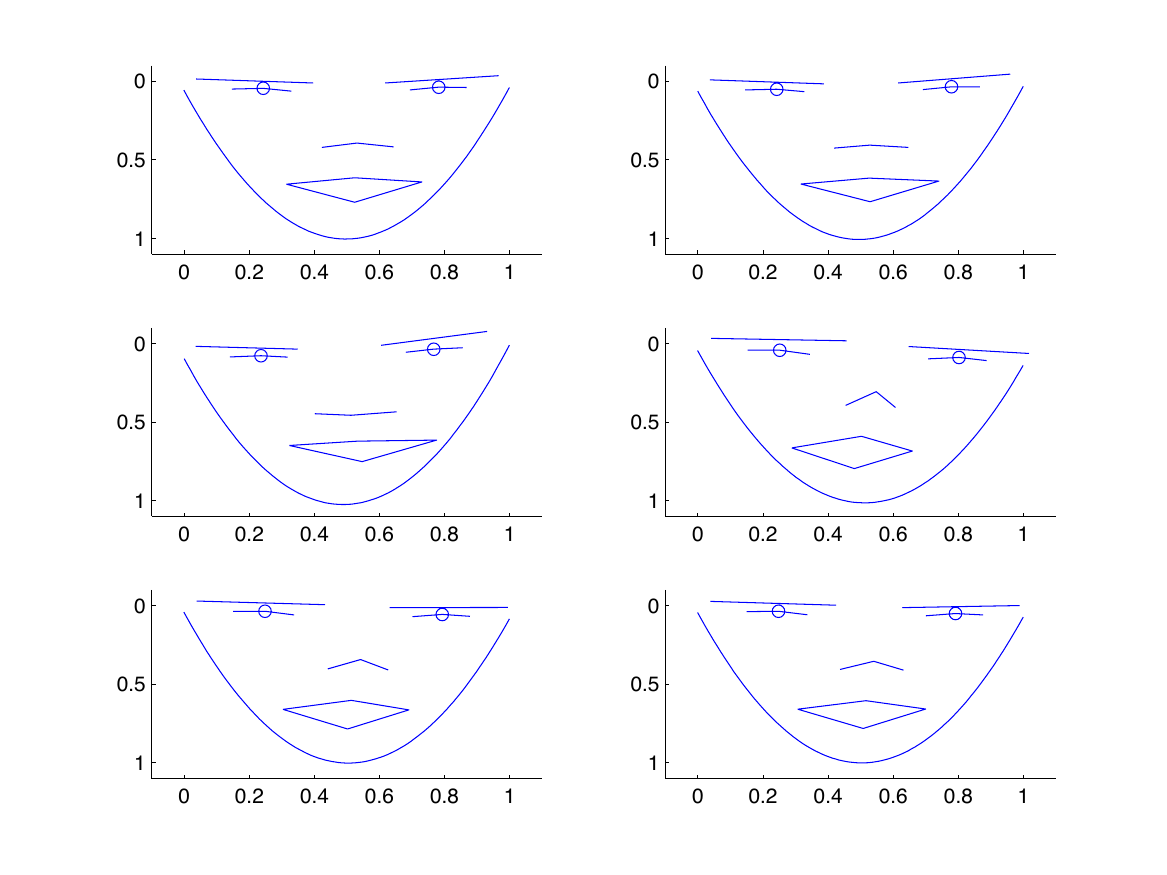}}
\caption{
The effect of varying the third Gaussian kernel PCA feature for ASM. 
}
\label{kPCA_ASM_3}
\end{center}
\vskip -0.2in
\end{figure}

For Gaussian kernel PCA, we set $\sigma=0.3275$ for the Gaussian kernel 
(the parameter selection method will be given in Section \ref{discussion}). 
To study the effects of each feature extracted with kernel PCA, we compute the mean 
$ \bar{y}_k $ and the standard deviation $ \sigma_{yk} $ of all the kernel PCA features. 
Each time, we uniformly sample $y_k$ in the range 
$ [ \bar{y}_k - c \sigma_{yk} , \bar{y}_k + c \sigma_{yk} ]$, where $c>0$ is a constant, and set 
$ y_{k'}=\bar{y}_{k'} $ for all  $k'\neq k$. When $c=3$, the effects of varying the first Gaussian kernel PCA feature, the second Gaussian kernel PCA feature, and the third Gaussian kernel PCA feature are shown in Figure \ref{kPCA_ASM_1}, Figure \ref{kPCA_ASM_2} and Figure \ref{kPCA_ASM_3}, respectively. 

By observation, we can see that the first PCA feature affects the orientation of the human face (left or right), and the second PCA feature to some extent determines some microexpression from amazement to calmness of the human face. In contrast, the Gaussian kernel PCA features seem to be determining some very different microexpressions than PCA features. 

%%%%%%%%%%%%%%%%%%%%%%%%%%%%%%%%%%%%%%%%%%%%%%%%%%%%%%%%%
%%%%%%%%%%%%%%%%%%%%%%%%%%%%%%%%%%%%%%%%%%%%%%%%%%%%%%%%%
%%%%%%%%%%%%%%%%%%%%%%%%%%%%%%%%%%%%%%%%%%%%%%%%%%%%%%%%%
\section{Discussions}
\label{discussion}
In this section, we address two concerns: First, how do we select the parameters 
for Gaussian kernel PCA; Second, what is the intuitive explanation of Gaussian kernel PCA. 
%%%%%%%%%%%%%%%%%%%%%%%%%%%%%%%%%%%%%%%%%%%%%%%%%%%%%%%%%
%%%%%%%%%%%%%%%%%%%%%%%%%%%%%%%%%%%%%%%%%%%%%%%%%%%%%%%%%
%%%%%%%%%%%%%%%%%%%%%%%%%%%%%%%%%%%%%%%%%%%%%%%%%%%%%%%%%
\subsection{Parameter Selection}

Parameter selection for kernel PCA directly determines the performance of the algorithm. For Gaussian kernel PCA, the most important parameter is the $\sigma$ in the kernel function defined by Eq. (\ref{eq30}). The Gaussian kernel is a function of the distance $ \Vert \mathbf{x-y} \Vert$ between two vectors $\mathbf{x}$ and $\mathbf{y}$. Ideally, if we want to separate different classes in the new feature space, then the parameter $\sigma$ shoud be smaller than inter-class distances, and larger than inner-class distances. However, we don't know how many classes are there in the data, thus it is not easy to estimate the inter-class or inner class distances. Alternatively, we can set $\sigma$ to a small value to capture only the neighborhood information of each data point. For this purpose, for each data point $\mathbf{x}_i$, let the distance from $\mathbf{x}_i$ to its nearest neighbor be $d_i^\mathrm{NN}$. In our experiments, we use this parameter selection strategy:
\begin{eqnarray}
\sigma=5\cdot\mean\limits_i\,(d_i^\mathrm{NN}) .
\end{eqnarray}
This strategy, in our experiments, ensures that the $\sigma$ is large enough to capture neighboring data points, and is much smaller than inter-class distances. When using different datasets, this strategy may need modifications. 

For the pre-image reconstruction of Gaussian kernel PCA, the initial guess $\mathbf{z}_0$ will determine whether the iterative algorithm (\ref{eq36}) converges. We can always simply use the mean of the training data as the initial guess: 
\begin{equation}
\mathbf{z}_0=
\mean \limits _i \, ( \mathbf{x}_i ) . 
\end{equation}

%%%%%%%%%%%%%%%%%%%%%%%%%%%%%%%%%%%%%%%%%%%%%%%%%%%%%%%%%
%%%%%%%%%%%%%%%%%%%%%%%%%%%%%%%%%%%%%%%%%%%%%%%%%%%%%%%%%
%%%%%%%%%%%%%%%%%%%%%%%%%%%%%%%%%%%%%%%%%%%%%%%%%%%%%%%%%
\subsection{Intuitive Explanation of Gaussian Kernel PCA}
We can see that in our synthetic data classification experiment, Gaussian kernel PCA with a properly selected parameter $\sigma$ can perfectly separate the two classes in an unsupervised manner, which is impossible for standard PCA. In the human face images classification experiment, Gaussian kernel PCA has a lower training error rate and a much lower testing error rate than standard PCA. From these two experiments, we can see that Gaussian kernel PCA reveals more complex hidden structures of the data than standard PCA. An intuitive understanding of the Gaussian kernel PCA is that it makes use of the distances between different training data points, which is like $k$-nearest neighbor or clustering methods. With a well selected $\sigma$, Gaussian kernel PCA will have a proper capture range, which will enhance the connection between data points that are close to each other in the original feature space. Then by applying eigenvector analysis, the eigenvectors will describe the directions in a high-dimensional space in which different clusters of data are scattered to the greatest extent. 

In this paper we are mostly using Gaussian kernel for kernel PCA. This is because it is intuitive, easy to implement, and possible to reconstruct the pre-images. However, we indicate that there are techniques to find more powerful kernel matrices by learning-based methods \cite{kPCA2, SDP_kPCA}.

%%%%%%%%%%%%%%%%%%%%%%%%%%%%%%%%%%%%%%%%%%%%%%%%%%%%%%%%%
%%%%%%%%%%%%%%%%%%%%%%%%%%%%%%%%%%%%%%%%%%%%%%%%%%%%%%%%%
%%%%%%%%%%%%%%%%%%%%%%%%%%%%%%%%%%%%%%%%%%%%%%%%%%%%%%%%%
\section{Conclusion}

In this paper, we discussed the theories of PCA, kernel PCA and ASMs. Then we 
focused on the pre-image reconstruction for Gaussian kernel PCA, and used this technique to design kernel PCA-based ASMs. We tested kernel PCA dimensionality reduction on synthetic data and human face images, and found that Gaussian kernel PCA succeeded in revealing more complicated structures of data than standard PCA, and achieving much lower classification error rates. We also implemented the Gaussian kernel PCA-based ASM and tested it on human face images. We found that Gaussian kernel PCA-based ASMs are promising in providing more deformation patterns than traditional ASMs. A potential application is that we could combine traditional ASMs and Gaussian kernel PCA-based ASMs for microexpression recognition on human face images. Besides, we proposed a parameter selection method to find the proper parameters for Gaussian kernel PCA, which works well in our experiments. 

\section*{Acknowledgment}
The author would like to thank Johan Van Horebeek (horebeek@cimat.mx) and Flor Martinez (flower10@cimat.mx) from CIMAT for reviewing our code and reporting the bugs. 

\nocite{non_d_r}

\bibliography{example_paper}
\bibliographystyle{icml2011}

\end{document}